\documentclass[11pt]{article}
\usepackage[preprint]{acl}

\usepackage{times}
\usepackage{latexsym}

\usepackage[T1]{fontenc}

\usepackage[utf8]{inputenc}

\usepackage{microtype}

\usepackage{inconsolata}

\usepackage{graphicx}

\usepackage{multirow}
\usepackage{tabularx}
\usepackage{float}
\usepackage{caption}
\usepackage{graphicx}%
\usepackage{multirow}%
\usepackage{amsmath,amssymb,amsfonts}%
\usepackage{amsthm}%
\usepackage{mathrsfs}%
\usepackage[title]{appendix}%
\usepackage{textcomp}%
\usepackage{manyfoot}%
\usepackage{booktabs}%
\usepackage{algorithm}%
\usepackage{algorithmicx}%
\usepackage{algpseudocode}%
\usepackage{listings}%
\usepackage{multicol}
\usepackage{makecell}
\usepackage{subcaption}
\usepackage{enumitem}

\usepackage{bm}
\usepackage[table]{xcolor}
\definecolor{darkred}{rgb}{0.55, 0.0, 0.0}  
\definecolor{darkgreen}{rgb}{0.0, 0.5, 0.0}  
\definecolor{cvprblue}{rgb}{0.21,0.49,0.74}
\definecolor{myblue}{RGB}{220,230,242}  
\definecolor{mygreen}{RGB}{226,240,217} 
\definecolor{mygray}{gray}{0.9}
\title{Triggering Chain-of-Thought via Latent Feature Interventions in Large Language Models}

\author{
 \textbf{Zhenghao He},
 \textbf{Guangzhi Xiong},
 \textbf{Bohan Liu},
 \textbf{Sanchit Sinha},
 \textbf{Aidong Zhang}
\\
 University of Virginia, Virginia, USA
 \\
    \texttt{\{zhenghao, guangzhi, qzp4ta, sanchit, aidong\}@virginia.edu}
}

\begin{document}
\maketitle
\begin{abstract}
Chain-of-Thought (CoT) prompting often improves the reasoning performance of large language models (LLMs), but the internal signal that triggers this behavior remains poorly understood.
Leveraging the sparse features captured by Sparse Autoencoders (SAEs), we propose a systematic framework to analyze and intervene on the internal representations of LLMs, identifying a small set of latent features that are linked to reasoning behavior and can be causally tested through targeted intervention.
Across multiple model families and reasoning benchmarks, we show that steering one or a small number of reasoning-related latent features can substantially induce reasoning behavior without explicit CoT prompting, achieving accuracy comparable to CoT.
We further show that the identified features are not tied to particular wording patterns or verbosity, and confirm their causal role in reasoning through suppression experiments that impair performance even under CoT prompting. These results suggest that CoT prompting activates specific latent features to trigger reasoning, and that targeted intervention on these features offers an alternative pathway to elicit efficient reasoning behavior without explicit CoT prompting.
Code is available at \url{https://github.com/Zhenghao-He/LatentCoT}.

\end{abstract}

\section{Introduction}
\begin{figure}[h]
    \centering
    \includegraphics[width=0.95\linewidth]{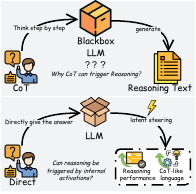}
    \caption{
    \textbf{Multiple triggers for latent reasoning in LLMs.}
    \textit{Top:} CoT prompting produces explicit reasoning text, while the internal mechanism responsible for its effectiveness remains unclear.
    \textit{Bottom:} We view reasoning as a latent internal mechanism that can be activated through different triggers, including latent steering.
    }
    \label{fig:intro}
\end{figure}
Chain-of-Thought (CoT) prompting has become a standard technique for improving reasoning in large language models, with consistent gains across arithmetic~\cite{lewkowycz2022solving}, symbolic, and logical reasoning tasks~\cite{wei2022chain, kojima2022large, talmor2019commonsenseqa, wang2022self}.
Yet a fundamental question remains open: \emph{is CoT prompting the unique mechanism that enables multi-step reasoning, or is it merely one of several ways to activate a latent reasoning capability?}
If reasoning corresponds to an internal computational state, it should in principle be possible to activate this state through means other than explicit step-by-step prompting (Figure~\ref{fig:intro}).

Recent studies provide preliminary evidence for this view.
Reasoning-relevant trajectories can be elicited by modifying the decoding process without explicit prompting~\cite{wang2024chain}, and continuous ``soft thought'' representations can enhance reasoning without generating step-by-step text~\cite{xu2025softcot}.
Moreover, causal analyses suggest that CoT traces are not necessarily the mechanism that produces the final answer~\cite{bao2024llms}.
However, these approaches either operate at the output level or lack direct causal evidence linking specific internal representations to reasoning behavior.


In this work, we analyze and intervene on the internal representations of large language models to investigate this question.
A natural starting point is to steer the model along the mean activation difference between CoT and direct prompting, following standard representation engineering approaches~\cite{tang2025unlockinggenerallongchainofthought,ma2025cotvalvelengthcompressiblechainofthoughttuning,sheng2026reasoningstrengthplanninglarge,lin2025controllingthinkingspeedreasoning,venhoff2025basemodelsknowreason}.
However, such dense steering directions entangle reasoning-relevant signals with formatting, verbosity, and other prompt-dependent artifacts, limiting both their effectiveness and cross-task generalization.
To address this, we employ Sparse Autoencoders (SAEs) to decompose model activations into disentangled latent features, enabling targeted intervention on dimensions associated with reasoning.
This design choice is empirically validated: across all models tested, SAE-based sparse feature steering substantially outperforms dense steering, and in some cases dense steering fails entirely while our method produces large accuracy gains (Section~\ref{sec:dense}).

Using this approach across three model families and six models, we find that steering one or a small number of latent features at the first generation step can induce coherent reasoning behavior under direct prompting. 
We further show that the identified features are not tied to particular wording patterns or formatting (Section~\ref{sec:cross_prompt}),
and that these features are causally involved in reasoning, as suppressing them substantially impairs performance even under CoT prompting (Section~\ref{sec:anti-steer}).
Together, these results suggest that CoT prompting activates specific latent features to trigger reasoning, and that targeted intervention on these features offers an alternative pathway to elicit reasoning without explicit CoT prompting.
 
Our main contributions are:
\begin{itemize}[leftmargin=*,itemsep=0pt,topsep=0pt]
    \item We propose a two-stage framework using SAEs to identify latent features associated with reasoning, and show that SAE-based steering substantially outperforms dense steering, highlighting the necessity of disentangled representations.
    \item Across six models from three families (up to 70B) and three benchmarks, we show that steering the identified features induces reasoning behavior under direct prompting, achieving accuracy comparable to CoT while producing shorter outputs.
    \item Through multiple controlled experiments, we provide evidence that the identified features are causally linked to reasoning computation, generalize across tasks and prompting styles, and are not reducible to surface formatting patterns.
\end{itemize}

\section{Related Work}

\noindent\textbf{Chain-of-Thought and Multi-step Reasoning.}
Chain-of-thought (CoT) prompting has been widely adopted as a practical approach for improving performance on tasks that require multi-step reasoning \citep{wei2022chain}. Prior work has demonstrated that encouraging models to produce intermediate reasoning steps can lead to substantial gains across a variety of domains, including arithmetic problem solving \citep{lewkowycz2022solving}, symbolic manipulation, and logical inference \citep{talmor2019commonsenseqa}. As a result, CoT-style prompting has become a common component in reasoning benchmarks and evaluation protocols for large language models, and has inspired numerous extensions such as self-consistency \citep{wang2022self}, structured reasoning prompts \citep{kojima2022large}, and methods that improve reasoning
through attention-guided adaptation \citep{sinha2026attentionguidedfinetuningmultimodallarge}.

\noindent\textbf{Reasoning Beyond Explicit CoT Prompting.}
Beyond the standard CoT prompting paradigm, a growing line of work suggests that multi-step reasoning behavior in large language models need not be uniquely tied to explicit CoT prompts. For instance, recent studies show that CoT-style reasoning trajectories can be elicited by altering the decoding process without using explicit prompting~\cite{wang2024chain}, and that implicit reasoning leveraging internal hidden states can support complex reasoning without generating step-by-step text~\cite{deng2023implicit}. Moreover, methods based on continuous or latent representations, such as soft thought tokens, demonstrate enhanced reasoning capability without relying on explicit verbal reasoning steps~\cite{xu2025softcot}. Complementary empirical analyses further indicate that the effectiveness of CoT prompting does not strictly depend on correct or valid intermediate chains, suggesting that the internal drives for reasoning extend beyond the surface verbal structure~\cite{wang2023towards}. 

\begin{figure*}[t]
    \centering
    \includegraphics[width=0.9\textwidth]{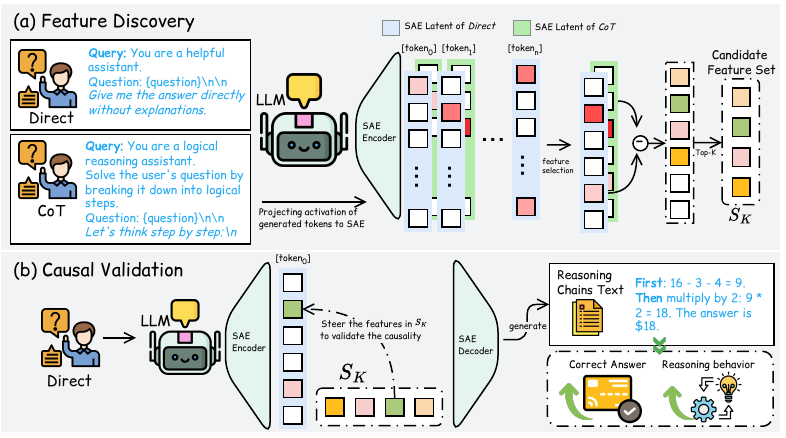}
    \caption{\textbf{Overview of the proposed two-stage pipeline.}
(a) \emph{Feature Discovery.} We contrast direct and chain-of-thought prompting to extract token-level activations, project them into sparse latent features using a pretrained sparse autoencoder (SAE), and identify prompt-sensitive candidate features via differential analysis.
(b) \emph{Causal Validation.} We apply targeted latent steering to selected features and inject the resulting residual into the model to assess their intervention sensitivity, evaluating the effect on model behavior and answer correctness.}
    \label{fig:pipline}
\end{figure*}

\noindent\textbf{Internal Representations and Activation Steering for Reasoning.}
Beyond output-level analyses, prior work studies reasoning through internal
representations. Probing, activation analysis, causal tracing, and concept-based
representation analysis associate hidden states with meaningful
behaviors and concepts~\cite{burns2212discovering,meng2022locating,he2025gcav}.
In reasoning settings, such analyses suggest that substantial computation can
occur internally even without explicit verbalization.

Activation-steering methods provide a complementary approach for controlling
model behavior through internal representations. Methods such as ActAdd
\cite{turner2024steering} and Contrastive Activation Addition (CAA)
\cite{rimsky2024steering} construct dense steering directions from contrastive prompts
or examples. While effective for behavioral control, such directions may
entangle the target behavior with other activation differences induced by the
contrastive conditions.

Sparse autoencoders (SAEs) instead decompose model activations into sparse
latent features, enabling more localized and interpretable interventions
\cite{cunningham2023sparse}. Sparse latent representations and dictionary-learning approaches have been used for concept discovery, interpretation, and intervention across language and vision models
\cite{xiong2026toward,liu2026robustnesstypographicattacktrainingfree,he2026casl}, including in
reasoning-related settings \cite{wang2025improving}.
\citet{galichin2026have} identify reasoning-related SAE features in reasoning
models and examine their steering effects.
\citet{fang2026controllable} use SAE features to control fine-grained
reasoning strategies such as backtracking and cross-verification, while
\citet{zhang2025fantastic} discover behaviors such as reflection and
backtracking from step-level reasoning activations.

These studies primarily analyze or control representations within an
already-active reasoning process. In contrast, we study whether sparse latent
features can causally trigger reasoning under direct prompting, before any CoT
trace is generated.
\section{Method}
\label{sec:method}
In this section, we present a two-stage pipeline for identifying and intervening on latent features associated with reasoning-related behaviors in large language models. Chain-of-Thought prompting is used as a contrasting condition to reveal prompt-dependent differences in latent activations. As illustrated in Figure~\ref{fig:pipline}, we first extract sparse latent features using a pretrained sparse autoencoder (SAE) and identify candidate features (Section~\ref{sec:feature_discovery}). We then define a latent steering procedure to estimate the intervention sensitivity of individual features on training data, and evaluate the selected features on a held-out test set (Section~\ref{sec:causal_validation}).


\subsection{Problem Setup}
\label{sec:problem_setup}

We consider a pretrained large language model $f_\theta$ with fixed parameters. Given an input question $x$ and a prompting condition $p$, the model generates an output sequence autoregressively. We focus on two prompting regimes: direct prompting $p^{\text{dir}}$ and chain-of-thought prompting $p^{\text{CoT}}$, which differ only in their instructions:
\begin{equation}
y \sim f_\theta(x, p), \quad p \in \{p^{\text{dir}}, p^{\text{CoT}}\}.
\end{equation}

Let $h^{(p)}_{l,t} \in \mathbb{R}^d$ denote the hidden activation at layer $l$ and token position $t$ under prompting condition $p$, where
\begin{equation}
h_{l,t}^{(p)} = f_\theta^{(l)}(x, p, y_{<t}).
\end{equation}
Our analysis focuses on latent activations aggregated from the generation process, with the specific aggregation strategy selected empirically and described in later sections.

\subsection{Latent Feature Extraction with SAE}
\label{sec:sae_extraction}

To obtain a sparse and intervention-friendly latent representation of model activations, we employ a pretrained SAE. For a given hidden activation $h_{l,t}^{(p)} \in \mathbb{R}^d$ at layer $l$ and token position $t$, the SAE encoder maps it to a latent vector $z_{l,t}^{(p)} \in \mathbb{R}^m$:
\begin{equation}
z_{l,t}^{(p)} = \phi(W_{\text{enc}} h_{l,t}^{(p)} + b_{\text{enc}}),
\end{equation}
where $\phi(\cdot)$ denotes an element-wise activation function, depending on the specific SAE variant. The corresponding decoder reconstructs the activation as
\begin{equation}
\hat{h}_{l,t}^{(p)} = W_{\text{dec}} z_{l,t}^{(p)} + b_{\text{dec}}.
\end{equation}

The SAE is trained to minimize a reconstruction objective of the form
\begin{equation}
\mathcal{L}_{\text{SAE}} = 
\mathbb{E}\!\left[\left\lVert h_{l,t}^{(p)} - \hat{h}_{l,t}^{(p)} \right\rVert_2^2\right]
+ \lambda \, \Omega\!\left(z_{l,t}^{(p)}\right),
\end{equation}
where $\Omega(\cdot)$ denotes a sparsity-inducing regularizer on the latent activations, such as an $\ell_1$ penalty or a top-$k$ masking constraint, depending on the SAE variant. In this work, we use a fixed, pretrained SAE and keep its parameters frozen during the analysis.

\subsection{Differential Feature Discovery}
\label{sec:feature_discovery}

We identify candidate latent features by contrasting their activation statistics under different prompting regimes. For a fixed input distribution over questions $x \sim \mathcal{D}$, direct prompting and chain-of-thought prompting induce two conditional distributions over latent activations.

\noindent \textbf{Feature Aggregation.} Given a single generated sequence under the prompting condition $p$, let $\{z_{l,t}^{(p)}(x)\}_{t=1}^T$ denote the SAE latent activations extracted at layer $l$ across token positions. We first apply an aggregation function $g(\cdot)$ to obtain a fixed-dimensional representation for each example:
\begin{equation}
\tilde{z}^{(p)}(x) = g\!\left(\{z_{l,t}^{(p)}(x)\}_{t=1}^T\right),
\end{equation}
where $g(\cdot)$ denotes a fixed aggregation over token-level activations. In our main experiments, we select the latent representation at the first generation step. The rationale for first-step intervention is analyzed in Appendix~\ref{sec:steer_timing}. We also evaluated alternative aggregation strategies, including average/max pooling over the generated sequence, but found them to be consistently less effective.

We hypothesize that latent features associated with CoT-style behavior are activated early in generation, before the model converges toward the final answer. Aggregating over later tokens may therefore dilute these signals, consistent with prior findings that reasoning-related activations emerge early in the generation process~\cite{david2025temporal}.

\noindent \textbf{Candidate Feature Set. }We then define the conditional mean activation of latent feature $k$ under prompting condition $p$ as the empirical average of its aggregated activation across a dataset of inputs:
\begin{equation}
\mu_k(p) =
\mathbb{E}_{x \sim \mathcal{D}}\!\left[\tilde{z}_k^{(p)}(x)\right]
=
\frac{1}{|\mathcal{D}|}
\sum_{x \in \mathcal{D}} \tilde{z}_k^{(p)}(x).
\end{equation}

Using these statistics, we compute a feature-wise differential score that captures prompt-induced modulation:
\begin{equation}
\Delta z_k = \mu_k(p^{\text{CoT}}) - \mu_k(p^{\text{dir}}).
\end{equation}

We select latent features that are most sensitive to changes in the prompting strategy by choosing the $K$ dimensions with the largest absolute differential scores. Formally, we define the selected feature set as
\begin{equation}
S_K = \arg\max_{S \subseteq \{1,\dots,m\},\, |S| = K}
\sum_{k \in S} \left| \Delta z_k \right|.
\end{equation}
This procedure identifies a compact set of prompt-sensitive latent features, without introducing additional modeling assumptions or learned probes.

\begin{table*}[ht]
\centering
\renewcommand{\arraystretch}{0.8}
\newcommand{\pmgray}[1]{\textcolor{gray}{\text{\scriptsize$\pm$#1}}}
\resizebox{\textwidth}{!}{
\begin{tabular}{c|l|cccccc}
\toprule
\multirow{2}{*}{\textbf{Model}} &
\multirow{2}{*}{\textbf{Strategy}} &
\multicolumn{2}{c}{\makecell[c]{\textbf{GSM8K} \\ \small \cite{cobbe2021gsm8k}}} &
\multicolumn{2}{c}{\makecell[c]{\textbf{GPQA} \\ \small \cite{rein2024gpqa}}} &
\multicolumn{2}{c}{\makecell[c]{\textbf{BBH} \\ \small \cite{suzgun2022challenging}}} \\
\cmidrule(lr){3-8} 
& & 
\textbf{Acc.}~($\uparrow$)&
\textbf{\#Tok}~($\downarrow$)
& 
\textbf{Acc.}~($\uparrow$)&
\textbf{\#Tok}~($\downarrow$)
& \textbf{Acc.}~($\uparrow$)&
\textbf{\#Tok}~($\downarrow$)
\\
\midrule
\multirow{4}{*}{\makecell[c]{LLaMA-3.1-8B-Instruct\\ \small \cite{meta_llama3_2024}}}
& Direct & 24.5 & 17\pmgray{34} & 28.7 & 2\pmgray{0}&50.8&3\pmgray{0}
\\
& \cellcolor{mygray}Steered Direct 
& \cellcolor{mygray}73.3 
& \cellcolor{mygray}83 \pmgray{53} 
& \cellcolor{mygray}28.2
& \cellcolor{mygray}43\pmgray{103}
& \cellcolor{mygray}61.6
& \cellcolor{mygray}53\pmgray{29}
 \\
& CoT  & 79.3 & 234\pmgray{70} & 20.7 & 423\pmgray{98}&77.2&146\pmgray{28}
\\
& \cellcolor{mygray}Steered CoT 
& \cellcolor{mygray}84.1 
& \cellcolor{mygray}227\pmgray{68} 
& \cellcolor{mygray}22.2
& \cellcolor{mygray}414\pmgray{99}
& \cellcolor{mygray}78.4
& \cellcolor{mygray}146\pmgray{27}
\\
\midrule
\multirow{4}{*}{\makecell[c]{LLaMA-3.3-70B-Instruct\\ \small \cite{llama3.3-70b-instruct}}}
& Direct & 46.7 & 12\pmgray{22} &41.9& 2\pmgray{0}&94.0&7\pmgray{7}
\\
& \cellcolor{mygray}Steered Direct 
& \cellcolor{mygray}88.8 
& \cellcolor{mygray}53\pmgray{30} 
& \cellcolor{mygray}44.9
& \cellcolor{mygray}47\pmgray{120}
& \cellcolor{mygray}93.6
& \cellcolor{mygray}40\pmgray{31}
\\
& CoT &96.1 & 268\pmgray{63}& 19.7 & 495\pmgray{41}&100.0&226\pmgray{44}
 \\
& \cellcolor{mygray}Steered CoT 
& \cellcolor{mygray}96.5 
& \cellcolor{mygray}257\pmgray{66} 
& \cellcolor{mygray}18.6
& \cellcolor{mygray}474\pmgray{66}
& \cellcolor{mygray}100.0
& \cellcolor{mygray}228\pmgray{39}
\\
\midrule
\multirow{4}{*}{\makecell[c]{Qwen3-0.6B\\ \small \cite{qwen3technicalreport}}}
& Direct &7.9& 14\pmgray{15}&26.7& 7 \pmgray{4}&39.6&13\pmgray{3}
\\
& \cellcolor{mygray}Steered Direct 
& \cellcolor{mygray}60.6 
& \cellcolor{mygray}357\pmgray{124} 
& \cellcolor{mygray}24.7
& \cellcolor{mygray}237\pmgray{224}
& \cellcolor{mygray}66.4
& \cellcolor{mygray}301\pmgray{309}
\\
& CoT  & 59.7 & 400\pmgray{108}&15.6&508\pmgray{18}&82.8&537\pmgray{431}
 \\
& \cellcolor{mygray}Steered CoT
& \cellcolor{mygray}59.6 
& \cellcolor{mygray}393\pmgray{109}
& \cellcolor{mygray}16.7
& \cellcolor{mygray}506\pmgray{24}
& \cellcolor{mygray}78.4
& \cellcolor{mygray}559\pmgray{487}
\\
\midrule
\multirow{4}{*}{\makecell[c]{Qwen3-4B\\ \small \cite{qwen3technicalreport}}}
& Direct & 34.5 & 32\pmgray{25} &32.8 &8\pmgray{5}&83.2&13\pmgray{2}
\\
& \cellcolor{mygray}Steered Direct 
& \cellcolor{mygray}60.0
& \cellcolor{mygray}141\pmgray{127}
& \cellcolor{mygray}27.7
& \cellcolor{mygray}418\pmgray{171}
& \cellcolor{mygray}88.4
& \cellcolor{mygray}74\pmgray{138}
\\
& CoT & 61.1 & 421\pmgray{101} &18.8&510\pmgray{12}&98.8&458\pmgray{328}
 \\
& \cellcolor{mygray}Steered CoT 
& \cellcolor{mygray}62.5
& \cellcolor{mygray}422\pmgray{101}
& \cellcolor{mygray}19.2
& \cellcolor{mygray}511\pmgray{10}
& \cellcolor{mygray}97.6
& \cellcolor{mygray}528\pmgray{348}
\\
\midrule
\multirow{4}{*}{\makecell[c]{Gemma-3-4B-Instruct\\ \small \cite{gemma_2025}}}
& Direct & 8.4 & 6\pmgray{17} &30.8 & 3\pmgray{0} &59.6&5\pmgray{0}
\\
& \cellcolor{mygray}Steered Direct  
& \cellcolor{mygray}74.0
& \cellcolor{mygray}243\pmgray{158} 
& \cellcolor{mygray}21.7
& \cellcolor{mygray}353\pmgray{188}
& \cellcolor{mygray}65.6
& \cellcolor{mygray}191\pmgray{203}
\\
& CoT &78.1&193\pmgray{92}&26.2&291\pmgray{219}&50.0&161\pmgray{155}
\\
& \cellcolor{mygray}Steered CoT
& \cellcolor{mygray}82.8
& \cellcolor{mygray}232\pmgray{104}
& \cellcolor{mygray}19.2
& \cellcolor{mygray}415\pmgray{147}
& \cellcolor{mygray}80.8
& \cellcolor{mygray}187\pmgray{122}
\\
\midrule
\multirow{4}{*}{\makecell[c]{Gemma-3-12B-Instruct\\ \small \cite{gemma_2025}}}
& Direct & 18.2 & 5\pmgray{2}&32.3&2\pmgray{0}&86.4&4\pmgray{0}
\\
& \cellcolor{mygray}Steered Direct
& \cellcolor{mygray}70.9
& \cellcolor{mygray}101\pmgray{98}
& \cellcolor{mygray}33.3
& \cellcolor{mygray}18\pmgray{26}
& \cellcolor{mygray}95.6
& \cellcolor{mygray}36\pmgray{18}
\\
& CoT & 92.7 & 181\pmgray{75}&24.4&404\pmgray{133}&95.2&136\pmgray{70}
 \\
& \cellcolor{mygray}Steered CoT
& \cellcolor{mygray}92.8 
& \cellcolor{mygray}168\pmgray{71}
& \cellcolor{mygray}22.2
& \cellcolor{mygray}400\pmgray{137}
& \cellcolor{mygray}96.8
& \cellcolor{mygray}120\pmgray{58}
\\
\bottomrule
\end{tabular}
}
\caption{
Reasoning performance across models and benchmarks under different strategies.
For each model, we compare direct prompt, Chain-of-Thought (CoT) prompting, and their steered variants.
Latent steering is applied only at the first generation step along an identified reasoning-related feature, with all subsequent decoding steps left unchanged.
Accuracy (\%) and average number of generated tokens (\#Tok) are reported for each setting.
}
\label{tab:statistic_result}
\end{table*}

\subsection{Causal Validation via Latent Steering}
\label{sec:causal_validation}

Given the candidate features $S_K$ identified in Section~\ref{sec:feature_discovery}, we define a latent steering procedure for a controlled intervention on model activations.

Let $a_{l,t}(x) \in \mathbb{R}^m$ denote the pre-activation latent representation produced by the SAE encoder at layer $l$ and generation step $t$. The corresponding post-activation latent is given by $z_{l,t}(x) = \phi(a_{l,t}(x))$. For a chosen intervention set $S \subseteq S_K$, we define an intervention by modifying the corresponding pre-activation dimensions:

{\small
\begin{equation}
\tilde{a}_{l,t,k}^{\text{steer}}(x; S) =
\begin{cases}
a_{l,t,k}(x) + \alpha \, \textstyle\mathbb{E}\!\left[|a_{l,t,k}(x)|\right], & k \in S, \\
a_{l,t,k}(x), & k \notin S,
\end{cases}
\end{equation}
}
where $\alpha$ is a scalar steering coefficient that controls the intervention strength.

We adopt an additive intervention on pre-activation latents to externally activate features, rather than amplify their current values, allowing the intervention to act even when a feature is inactive.
The perturbation is normalized by the feature’s empirical activation scale, ensuring comparable intervention strength across models.

The steered latent representation is obtained as
\begin{equation}
\tilde{z}_{l,t}^{\text{steer}}(x) = \phi\!\left(\tilde{a}_{l,t}^{\text{steer}}(x; S)\right).
\end{equation}

The modified latent representation $\tilde{z}_{l,t}^{\text{steer}}(x)$ is mapped back to the activation space using the SAE decoder,
\begin{equation}
\hat{h}_{l,t}^{\text{steer}}(x) = W_{\text{dec}} \tilde{z}_{l,t}^{\text{steer}}(x) + b_{\text{dec}}.
\end{equation}

Because the sparse autoencoder produces imperfect reconstruction, directly injecting the decoded steered representation may introduce reconstruction bias. Therefore, we apply a residual injection scheme. Let $\hat{h}_{l,t}(x) = \mathrm{Dec}(z_{l,t}(x))$ denote the reconstruction of the original activation. The activation injected into the model is given by
\begin{equation}
\tilde{h}_{l,t}(x) = h_{l,t}(x) + \left( \hat{h}_{l,t}^{\text{steer}}(x) - \hat{h}_{l,t}(x) \right).
\end{equation}
This residual correction ensures a localized intervention that isolates the effect of latent steering with minimal side effects.

Rather than intervening on all candidate features simultaneously, we estimate the intervention sensitivity of individual features on a held-out training set by restricting to singleton interventions $S=\{k\}$ for each $k \in S_K$. For each candidate feature, we apply the latent steering procedure and measure the resulting change. Features with consistently positive intervention sensitivity are selected.
The features are selected based on the training data and are fixed during evaluation.
\section{Experiments}

\subsection{Implementation Details}

For each model, we identify reasoning-related SAE features at a single 
mid-to-late transformer layer and perform steering at the same layer 
throughout all experiments. The steering layer, feature index, and 
strength $\alpha$ are selected on a validation split (Section~\ref{sec:causal_validation}), fixed per model, 
and used consistently across all datasets. Model-specific configurations 
are provided in Appendix~\ref{sec:config}.
Unless otherwise specified, accuracy is reported from a single evaluation run with fixed decoding settings, while generated-token statistics are reported as the mean and standard deviation across evaluation examples.

\noindent\textbf{Datasets.}
We evaluate our method on three reasoning benchmarks: GSM8K~\cite{cobbe2021gsm8k}, GPQA~\cite{rein2024gpqa}, and Big-Bench Hard (BBH)~\cite{suzgun2022challenging}.
For each base model, reasoning-related features are identified using 1,000 question–answer pairs sampled from the \emph{training split of GSM8K}.
All reported evaluation results are obtained by steering the identified features on the GSM8K test set, the GPQA Diamond subset, and the logical\_deduction\_three\_objects task from Big-Bench Hard (BBH).

\noindent\textbf{Models.}
We study six language models of varying sizes: LLaMA-3.1-8B-Instruct, LLaMA-3.3-70B-Instruct, Qwen-3-0.6B, Qwen-3-4B, Gemma-3-4B-Instruct, and Gemma-3-12B-Instruct.
All models are used in inference-only mode, with weights frozen throughout all experiments.

\noindent\textbf{Sparse Autoencoders.}
We employ pretrained SAEs released by Goodfire~\cite{goodfire_sae_llama3} for LLaMA models and Gemma Scope 2~\cite{McDougallGemmaS2} for Gemma models. For Qwen models, we train SAEs from scratch; training details are provided in Appendix~\ref{sec:train_sae}.

\begin{figure}[t]
    \centering
    \includegraphics[width=\linewidth]{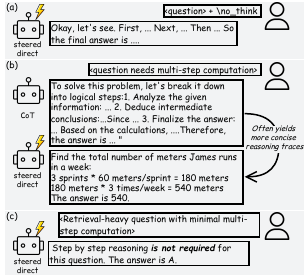}
    \caption{
    \textbf{Behaviors induced by steering.}
    (a) Steering can override prompt-level instructions that suppress reasoning (e.g., \texttt{\textbackslash no\_think}).
    (b) For questions requiring multi-step computation, steering can yield shorter explicit reasoning traces compared to standard chain-of-thought prompting.
    (c) On retrieval-heavy questions, latent steering leads the model to explicitly note that step-by-step reasoning is unnecessary.
    \emph{These examples are illustrative and not intended to be exhaustive.}
    }
    \label{fig:qualitative_examples}
\end{figure}

\subsection{Effects of Latent Steering on Reasoning Performance} 
For each base model, we first identify a small set of reasoning-related latent features using the procedure described in Section~\ref{sec:feature_discovery} and Section~\ref{sec:causal_validation}.
The specific feature indices selected for each model and the corresponding $\alpha$ are reported in Appendix~\ref{sec:steering_settings}.

During evaluation, we apply latent steering only at the \emph{first generation step} along the identified reasoning feature, following the step-wise intervention strategy introduced in Section~\ref{sec:causal_validation}.
All other model components, decoding parameters, and prompts are kept identical to the unsteered baselines.
The resulting performance is shown in Table~\ref{tab:statistic_result}.

\noindent\textbf{Steering improves reasoning accuracy on reasoning-intensive tasks.}
Compared to direct prompting, steered direct yields large accuracy gains on reasoning-intensive benchmarks such as GSM8K and BBH.
This indicates that the identified latent features are closely tied to internal reasoning processes.
Notably, steering can induce reasoning behavior even when the prompt explicitly discourages intermediate reasoning.
For example, on Qwen models, steered direct decoding produces coherent multi-step reasoning despite prompts containing the control token \texttt{\textbackslash no\_think}, which is designed to suppress the explicit generation of CoT (Figure~\ref{fig:qualitative_examples}a).
This behavior suggests that the intervention operates at the level of internal computation rather than surface prompt compliance.

\noindent\textbf{Early single-feature steering leads to more concise reasoning outputs in large models.}
For larger models such as LLaMA-3.3-70B, steering achieves accuracy comparable to or exceeding standard CoT prompting while substantially reducing the length of generated reasoning text (Figure~\ref{fig:qualitative_examples}b). Notably, this is achieved by steering a single latent feature (feature \#13709) at only the first generation step. 

\noindent\textbf{Steering provides limited benefit once the model is already in a reasoning mode.}
Steered CoT yields only marginal improvements over standard CoT.
We hypothesize that when the reasoning-related feature is already strongly activated under CoT prompting, additional amplification does not further enhance reasoning performance.
This observation is consistent with the view that steering primarily acts as a trigger for entering a reasoning mode, rather than refining an ongoing reasoning process, as further analyzed in Appendix~\ref{sec:reason_mode}.

\noindent\textbf{Limited gains on tasks with weak reliance on multi-step reasoning.}
On GPQA, where CoT prompting does not consistently outperform direct decoding, steering likewise provides little or no improvement.
In some cases, models refuse to do reasoning and directly produce answers when we strongly activate those reasoning features (Figure~\ref{fig:qualitative_examples}c).
This suggests that when a task does not strongly rely on multi-step reasoning, the identified features are less engaged and steering has limited effect. This is supported by Section~\ref{sec:cross_dataset}, where features identified from GPQA fail to improve GPQA itself but improve the other two benchmarks.

\subsection{Comparison with Dense Steering Baseline}
\label{sec:dense}
To compare our method with standard representation engineering approaches, we construct a dense CoT steering direction using the mean activation difference between CoT and direct prompting. Specifically, we extract the hidden state of the last prompt token at the same layer used for SAE feature discovery, and define the dense direction as the average difference between CoT and direct representations on the GSM8K training set.
During generation, the vector is added to the hidden state with a fixed steering strength $\alpha$, tuned on a validation split. Detailed implementation details are provided in Appendix~\ref{app:dense}.
The resulting direction is then evaluated on GSM8K test and BBH.

As shown in Table~\ref{tab:sae_vs_dense}, dense steering yields smaller improvements than SAE feature steering across all models. Moreover, the dense direction transfers poorly to BBH, whereas SAE steering consistently improves reasoning performance across both datasets. We also observe that dense steering is ineffective on Qwen3, where prompts include explicit \texttt{<no\_think>} tags.
One possible explanation is that the dense direction mainly captures dataset-specific activation differences between CoT and direct prompting. As a result, it may act as a task-dependent optimization direction and generalize poorly across tasks or prompting formats. In contrast, SAE isolates a sparse latent feature that more cleanly captures the reasoning signal, enabling more reliable intervention.

\begin{table}[t]
\centering
\renewcommand{\arraystretch}{0.9}
\newcommand{\pmgray}[1]{\textcolor{gray}{\text{\scriptsize$\pm$#1}}}
\resizebox{\columnwidth}{!}{
\begin{tabular}{c|l|cccc}
\toprule
\multirow{2}{*}{\textbf{Model}} &
\multirow{2}{*}{\textbf{Strategy}} &
\multicolumn{2}{c}{\makecell[c]{\textbf{GSM8K} \\ \small \cite{cobbe2021gsm8k}}} &
\multicolumn{2}{c}{\makecell[c]{\textbf{BBH} \\ \small \cite{suzgun2022challenging}}} \\
\cmidrule(lr){3-6} 
& & 
\textbf{Acc.}~($\uparrow$)&
\textbf{\#Tok}
& \textbf{Acc.}~($\uparrow$)&
\textbf{\#Tok}
\\
\midrule
\multirow{3}{*}{\makecell[c]{LLaMA-3.1\\-8B-Instruct\\ \small \cite{meta_llama3_2024}}}
& Direct & 24.5 & 17\pmgray{34} &50.8&3\pmgray{0}
\\
& Dense & \underline{68.4} & 88\pmgray{65} & \underline{56} & 3\pmgray{0} 
\\
& \cellcolor{mygray}SAE
& \cellcolor{mygray}\textbf{73.3 }
& \cellcolor{mygray}83 \pmgray{53} 
& \cellcolor{mygray}\textbf{61.6}
& \cellcolor{mygray}53\pmgray{29}
 \\
\midrule
\multirow{3}{*}{\makecell[c]{Qwen3-4B\\ \small \cite{qwen3technicalreport}}}
& Direct & 34.5 & 32\pmgray{25} &83.2&13\pmgray{2}
\\
& Dense & \underline{34.6}&32\pmgray{29} & \underline{83.6}&  13\pmgray{2}
\\
& \cellcolor{mygray}SAE 
& \cellcolor{mygray}\textbf{60.0}
& \cellcolor{mygray}141\pmgray{127}
& \cellcolor{mygray}\textbf{88.4}
& \cellcolor{mygray}74\pmgray{138}
\\
\midrule
\multirow{3}{*}{\makecell[c]{Gemma-3\\-4B-Instruct\\ \small \cite{gemma_2025}}}
& Direct & 8.4 & 6\pmgray{17} &59.6&5\pmgray{0}
\\
& Dense & \underline{16.2} &46\pmgray{118} &\underline{59.6} &  5\pmgray{0} 
\\
& \cellcolor{mygray}SAE  
& \cellcolor{mygray}\textbf{74.0}
& \cellcolor{mygray}243\pmgray{158} 
& \cellcolor{mygray}\textbf{65.6}
& \cellcolor{mygray}191\pmgray{203}
\\
\bottomrule
\end{tabular}
}
\caption{
Comparison between dense representation steering and sparse SAE feature steering under direct prompting. \textbf{Bold} indicates the best result for each model, while \underline{underlined} values denote the second-best results.
}
\label{tab:sae_vs_dense}
\end{table}


\subsection{Suppression under CoT Prompting}
\label{sec:anti-steer}

We further examine the causal role of the identified features by
suppressing their contributions during CoT prompting. Using Qwen3-4B,
we intervene on the top-25 identified reasoning-related features at
layer 29 throughout both prompt processing and generation. At each forward pass, we project the positive activations of the target SAE features back into the model hidden space using their corresponding
decoder directions, and subtract a scaled version of this contribution from the residual stream. The intervention magnitude is calibrated on held-out validation data for each dataset, while all other generation settings are kept unchanged.

As shown in Table~\ref{tab:suppression}, suppression substantially
reduces CoT accuracy on both GSM8K and BBH. This cross-dataset
degradation provides additional causal evidence that the identified
features contribute to reliable reasoning under CoT prompting.

\begin{table}[h]
\centering
\small
\begin{tabular}{llcc}
\toprule
\textbf{Dataset} & \textbf{Strategy}
& \textbf{Acc. (\%)}
& \textbf{Avg. Tokens} \\
\midrule
GSM8K & CoT         & 61.1 & 421 \\
      & CoT + Supp. & 12.1 & 479 \\
\midrule
BBH   & CoT         & 98.8 & 458 \\
      & CoT + Supp. & 62.8 & 401 \\
\midrule
GPQA  & CoT         & 18.8 & 510 \\
      & CoT + Supp. & 21.7 & 476 \\
\bottomrule
\end{tabular}
\caption{
Effect of suppressing the hidden-state contributions of the top-25
reasoning-related SAE features under CoT prompting using Qwen3-4B.
}
\label{tab:suppression}
\end{table}

Importantly, suppression does not cause generation length to collapse.
The suppressed outputs remain comparable in length to standard CoT
responses rather than reverting to short direct-answer generations.
Thus, the model can continue to produce extended responses while its
answer accuracy substantially deteriorates, suggesting that long-form
generation alone is not sufficient for reliable reasoning.

GPQA exhibits a different pattern: suppression slightly improves over the CoT baseline. This is consistent with our earlier observation that CoT prompting itself provides limited benefit on GPQA.

\subsection{Beyond Prompt Wording and Output Verbosity}
\label{sec:cross_prompt}

A natural concern is whether the identified features simply track specific prompt wording or output verbosity rather than reasoning itself. We address this from two complementary directions.

\begin{figure}[h]
    \centering
    \includegraphics[width=\linewidth]{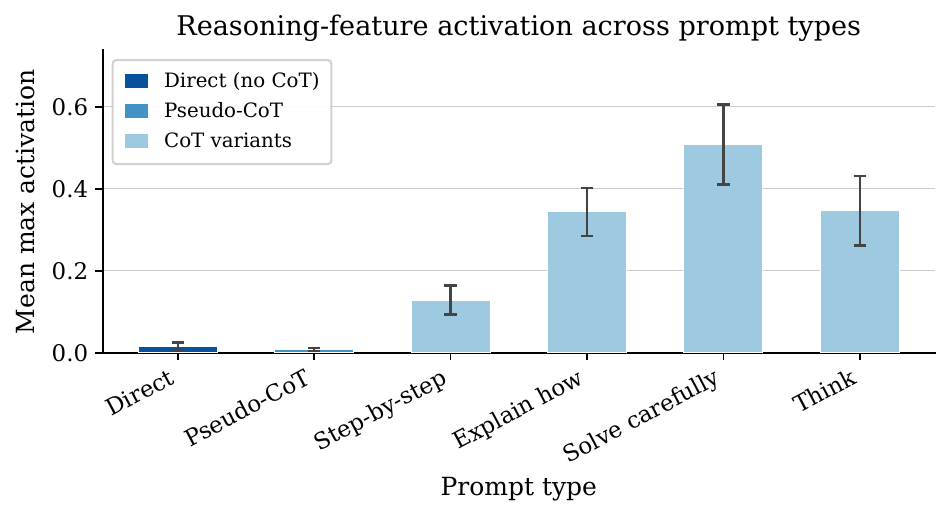}
    \caption{
    Activation of the reasoning-related SAE feature under different 
    prompting conditions on GSM8K using LLaMA-3.1-8B-Instruct. 
    Pseudo-CoT mimics step-by-step formatting without performing reasoning.
    }
    \label{fig:cross_prompt}
\end{figure}

\paragraph{Cross-prompt generalization.}
Figure~\ref{fig:cross_prompt} shows the activation of the identified 
reasoning feature under different prompting conditions on GSM8K. Several 
alternative prompts that encourage reasoning (e.g., ``Explain how'', 
``Solve carefully'', ``Think'') consistently activate the same feature, 
in some cases more strongly than explicit step-by-step prompting (detailed prompts can be found in Appendix~\ref{sec:input_prompts}). This indicates that the feature is not tied to a specific prompt template but responds to diverse linguistic cues that encourage reasoning.
Notably, stronger activation does not necessarily translate into higher
accuracy: despite exhibiting comparable or even stronger activation than
step-by-step prompting, these alternative prompts achieve similar
accuracy. This observation is consistent with our analysis in
Appendix~\ref{sec:reason_mode}, which suggests that activation of this
feature is better understood as a reasoning-mode signal than as a direct
predictor of answer correctness.

\paragraph{Pseudo-CoT.}
Different reasoning prompts also tend to produce similar discourse 
markers (e.g., ``First'', ``Next''). To rule out that the feature merely 
tracks such formatting patterns, we construct a \textit{pseudo-CoT} 
condition using the following prompt:
\begin{quote}
\small
``Provide a detailed restatement of the problem in a step-by-step format. 
Use expressions such as `First,' `Next,' and `Therefore,' but do not 
perform any calculations or intermediate reasoning. After the restatement, 
give the final answer directly.''
\end{quote}
As shown in Figure~\ref{fig:cross_prompt}, pseudo-CoT activates the 
feature even less than direct prompting, while CoT variants activate it 
substantially more strongly. Together, these results confirm that the 
feature is linked to reasoning computation, not surface formatting or 
verbosity.

\subsection{Feature Dynamics During Generation}

\begin{figure}[h]
    \centering
    \begin{subfigure}[b]{0.494\linewidth}
        \centering
        \includegraphics[width=\linewidth]{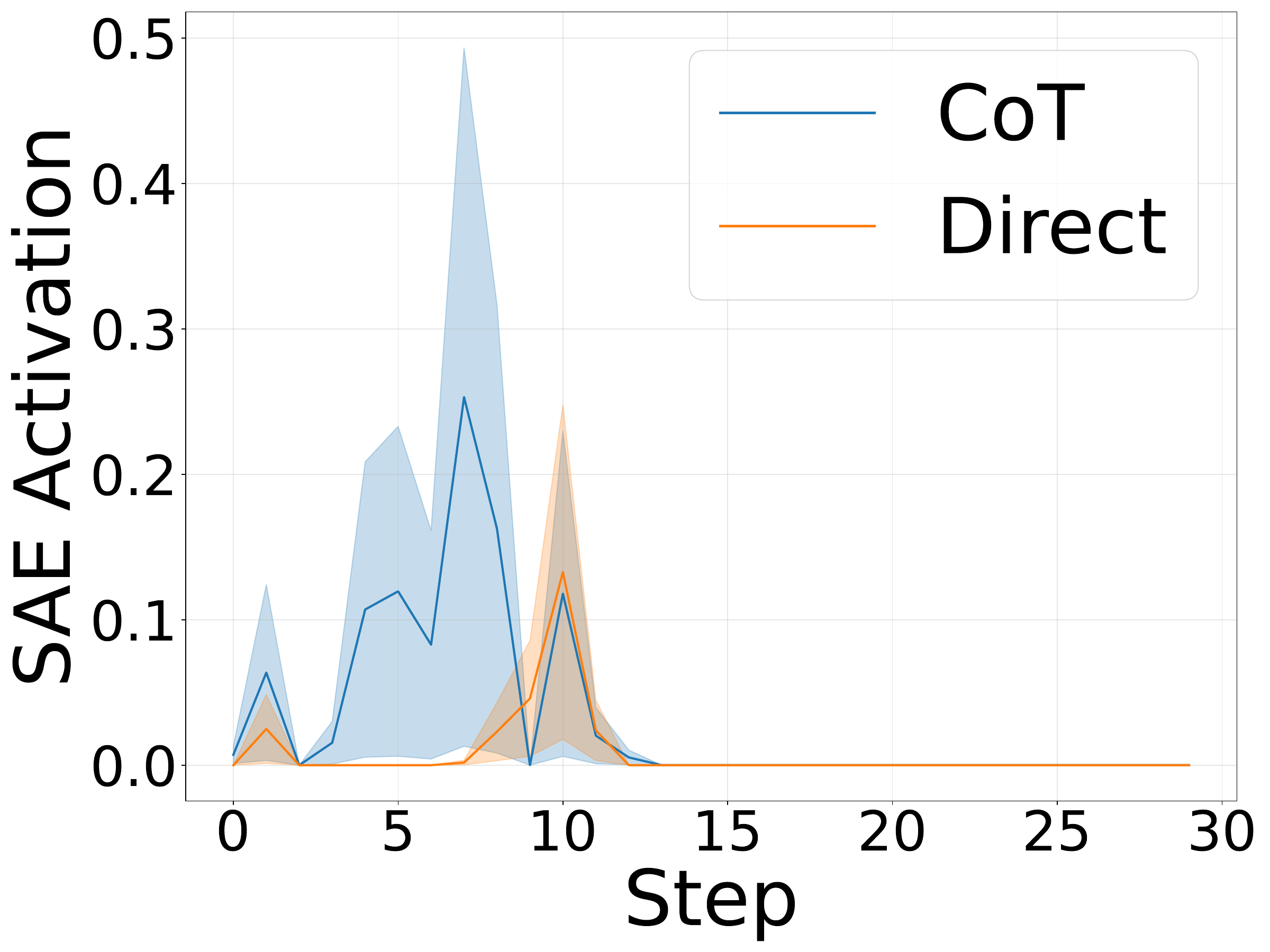}
        \caption{Reasoning feature \#8629 in LLaMA-3.1-8B-Instruct}
        \label{fig:cot}
    \end{subfigure}
    \hfill
    \begin{subfigure}[b]{0.494\linewidth}
        \centering
        \includegraphics[width=\linewidth]{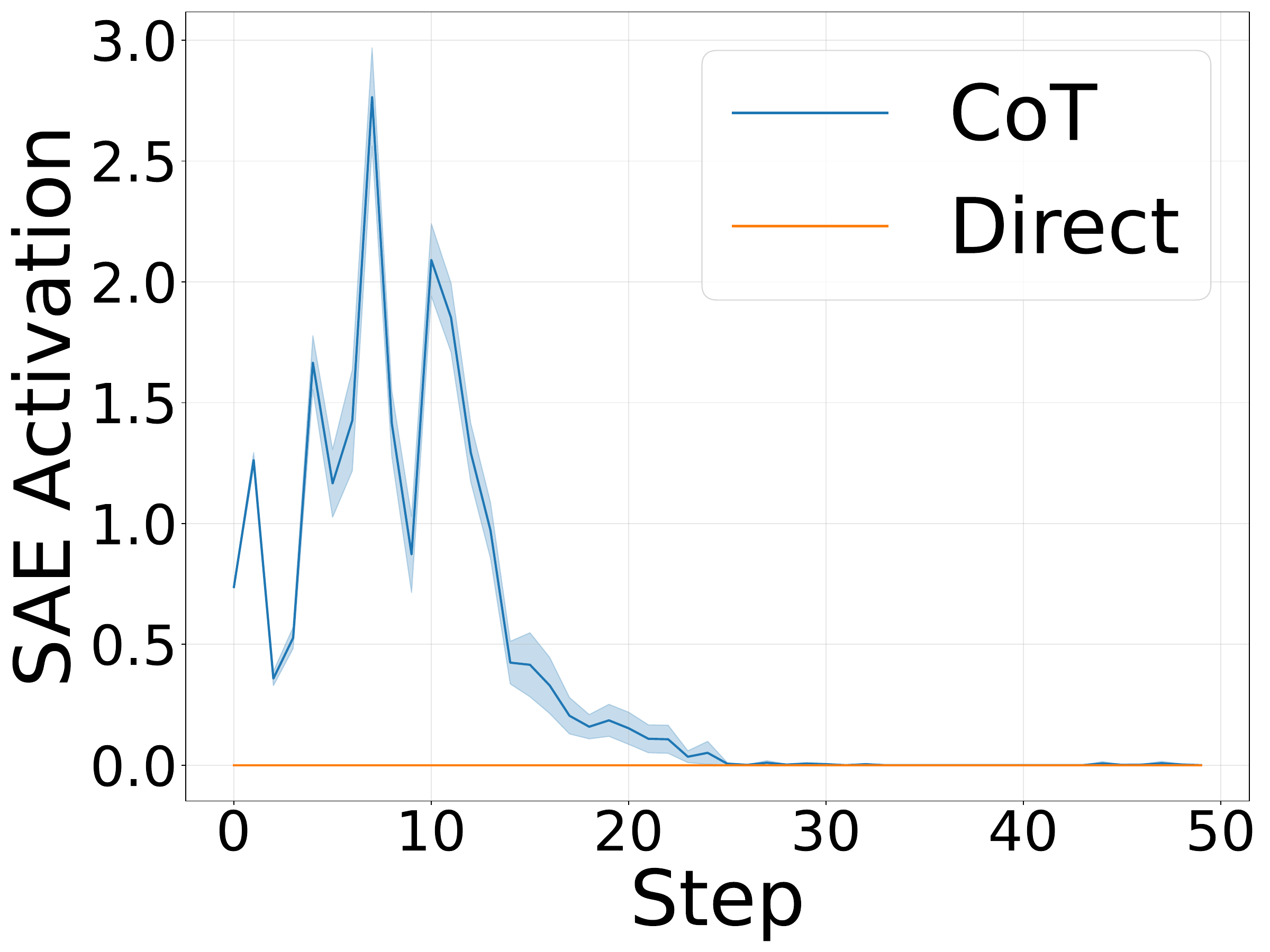}
        \caption{Reasoning feature \#13709 in LLaMA-3.3-70B-Instruct}
        \label{fig:direct}
    \end{subfigure}
    \caption{
    Activation dynamics of a reasoning-related SAE feature during generation on GSM8K.
    }
    \label{fig:dynamic}
\end{figure}

Figure~\ref{fig:dynamic} visualizes the activation dynamics of the reasoning-related SAE features used for steering during answer generation on GSM8K; these same single latent dimensions are intervened on to obtain all steered results reported in Table~\ref{tab:statistic_result}.
Across both models, the identified features exhibit a highly non-uniform temporal profile, with activations peaking early in decoding and rapidly decaying thereafter.
This transient pattern suggests that the feature is primarily engaged during the initial phase of reasoning rather than throughout generation.

Moreover, CoT prompting consistently induces stronger activation of the reasoning feature compared to direct decoding, indicating that the feature is associated with reasoning-oriented generation modes.
Despite differences in scale, this temporal structure is qualitatively consistent across model sizes, with larger models exhibiting more concentrated and higher-magnitude activation.
These observations support our focus on early interventions when steering reasoning-related latent features.

\subsection{Cross-Dataset Feature Identification}
\label{sec:cross_dataset}
To examine whether the identified feature is tied to a specific dataset or task, we perform cross-dataset feature identification and evaluation. Specifically, we independently extract reasoning-related features from GSM8K, GPQA, and BBH, and evaluate their steering effects across all three benchmarks.

\begin{figure}[h]
\centering
\includegraphics[width=0.75\linewidth]{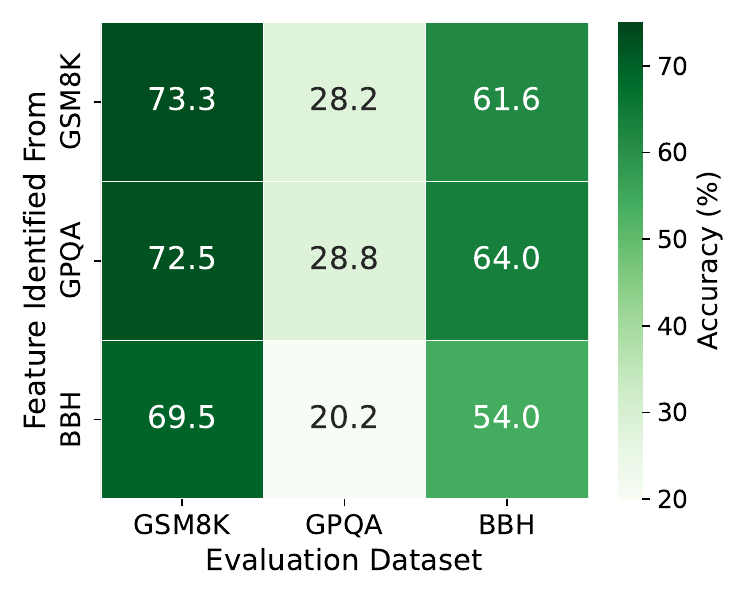}
\caption{
Cross-dataset evaluation of reasoning features identified from different datasets.
}
\label{fig:cross_dataset_heatmap}
\end{figure}

The results are shown in Figure~\ref{fig:cross_dataset_heatmap}. Features identified from GSM8K and GPQA both improve performance on GSM8K and BBH. Interestingly, the feature identified from GPQA does not improve GPQA itself, but still improves performance on GSM8K and BBH. This suggests that the identified feature is not limited to arithmetic reasoning. Instead, it indicates that activating CoT-style reasoning does not substantially benefit GPQA, while the reasoning feature remains useful on tasks where step-by-step reasoning improves performance.

To further characterize the extent to which feature discovery is shared
across datasets, we measure the overlap between the top-$k$ candidate
feature sets independently identified from each dataset. For two
datasets with top-$k$ feature sets $A_k$ and $B_k$, we compute

\begin{equation}
    \mathrm{Jaccard@}k(A,B)
    =
    \frac{|A_k \cap B_k|}
         {|A_k \cup B_k|}.
\end{equation}

We report the results for $k=25$ in
Table~\ref{tab:cross_dataset_overlap}. All dataset pairs exhibit
non-zero overlap, indicating that feature discovery recovers partially
shared reasoning-related SAE features rather than entirely
dataset-specific sets. The overlap is highest between GPQA and BBH
($0.28$), consistent with the strong GPQA-to-BBH transfer observed in
Figure~\ref{fig:cross_dataset_heatmap}. At the same time, the relatively limited
overlap across datasets indicates that feature identification remains
substantially dataset-dependent.

\begin{table}[t]
    \centering
    \renewcommand{\arraystretch}{0.9}
    \small
    \begin{tabular}{lccc}
        \toprule
        & GSM8K & GPQA & BBH \\
        \midrule
        GSM8K & 1.00 & 0.11 & 0.09 \\
        GPQA  & --   & 1.00 & \textbf{0.28} \\
        BBH   & --   & --   & 1.00 \\
        \bottomrule
    \end{tabular}
    \caption{
        Cross-dataset overlap of the top-25 candidate
        reasoning-related SAE features, measured by Jaccard similarity.
    }
    \label{tab:cross_dataset_overlap}
\end{table}
\section{Conclusion}


In this work, we show that multi-step reasoning in large language models can be triggered by intervening on latent features identified through sparse autoencoders. Across six models from three families, steering one or a small number of features induces reasoning behavior under direct prompting, approaching CoT accuracy with shorter outputs. We further show that these features are specific to reasoning computation rather than surface formatting, and suppressing them impairs performance under CoT prompting. These findings suggest that CoT prompting activates a latent reasoning mechanism that can also be triggered through targeted internal intervention, opening a pathway toward more efficient and controllable reasoning in language models.

\section*{Limitations}

Although our results are consistent with the interpretation that SAE representations partially disentangle latent features related to internal reasoning computation from those associated with verbose verbalization, we do not claim that the identified features constitute fully disentangled or isolated reasoning mechanisms. Our conclusions are based on behavioral and intervention-level evidence rather than complete mechanistic characterization. Our evaluation is limited to six models and three benchmarks, and broader validation across models and tasks remains for future work.

\section*{Acknowledgments}
This work is supported in part by the US National Science Foundation (NSF) under grants IIS-2106913, IIS-2538206, IIS-2529378, CCF-2217071, and CNS-2213700. Any recommendations expressed in this material are those of the authors and do not necessarily reflect the views of NSF. 
We also acknowledge the use of ChatGPT solely for assistance with grammar, punctuation, vocabulary, and language polishing.


\clearpage
\bibliography{custom}
\appendix
\clearpage
\newpage
\section{Appendix}
\label{sec:appendix}

\subsection{Dataset Details}

\subsubsection{GSM8K}
GSM8K is a benchmark of grade-school level mathematical word problems that require multi-step numerical reasoning~\cite{cobbe2021gsm8k}. Each example consists of a natural language question and a short numerical answer. Solving these problems typically involves parsing the problem description, performing a sequence of arithmetic operations, and producing a final numeric result. 

In our experiments, GSM8K is used both for identifying reasoning-related latent features and for evaluating reasoning performance. Specifically, we randomly sample 1,000 examples from the GSM8K training split to identify candidate reasoning-related features using the procedure described in Section~\ref{sec:feature_discovery}. All reported evaluation results are obtained on the GSM8K test split. Accuracy is measured as exact match between the model output and the ground-truth answer, following standard evaluation protocols.

\subsubsection{GPQA}
GPQA is a challenging question answering benchmark designed to assess scientific reasoning at the graduate level~\cite{rein2024gpqa}. The dataset covers multiple domains, including physics, chemistry, and biology, and is constructed to be resistant to surface-level retrieval by large language models. Questions often require combining domain knowledge with logical reasoning rather than simple factual recall.

We evaluate our method on the GPQA Diamond subset, which contains questions with verified difficulty and high annotation quality. Model performance is measured by exact match accuracy over multiple-choice answers. We note that prior work has shown that chain-of-thought prompting does not consistently improve performance on GPQA, making it a useful testbed for examining whether latent steering selectively benefits tasks that rely on multi-step reasoning.

\subsubsection{Big-Bench Hard}
Big-Bench Hard (BBH) is a curated subset of the BIG-Bench benchmark that focuses on tasks known to be challenging for language models~\cite{suzgun2022challenging}. The tasks in BBH are selected based on their low performance under standard prompting and often require logical deduction, symbolic manipulation, or multi-step reasoning.

In this work, we evaluate on the \texttt{logical\_deduction\_three\_objects} task from BBH, which requires reasoning about relational constraints among multiple entities. This task emphasizes structured logical reasoning rather than numerical computation. Model performance is evaluated using exact match accuracy on the final answer. BBH serves as a complementary benchmark to GSM8K, allowing us to test whether the identified reasoning-related features generalize beyond arithmetic reasoning to logical deduction tasks.

\subsection{Model Details}

We evaluate our method on six pretrained, instruction-tuned large language models drawn from three widely used model families: LLaMA-3, Qwen-3, and Gemma-3. These families differ in architecture, training data, and scale, allowing us to assess the generality of latent steering across diverse model designs. All models are used in inference-only mode, with parameters frozen throughout all experiments.

\subsubsection{LLaMA-3 Family}
We consider two instruction-tuned models from the LLaMA-3 family: LLaMA-3.1-8B-Instruct and LLaMA-3.3-70B-Instruct. These models represent small- and large-scale regimes within the same architectural family, enabling controlled comparisons across model size. LLaMA-3 models are widely adopted as strong open-weight baselines for reasoning tasks, and prior work has shown that chain-of-thought prompting is particularly effective for larger LLaMA models. We use pretrained sparse autoencoders released by Goodfire for both LLaMA-3 models.

\subsubsection{Qwen-3 Family}
We evaluate two models from the Qwen-3 family: Qwen-3-0.6B and Qwen-3-4B. These models provide additional coverage of smaller-scale language models and differ substantially from LLaMA-3 in training data composition and architectural choices. For Qwen-3 models, pretrained sparse autoencoders are not publicly available, so we train sparse autoencoders from scratch following the procedure described in Appendix~\ref{sec:train_sae}. Including Qwen-3 allows us to examine whether reasoning-related latent features and steering effects persist in models with more limited capacity.

\subsubsection{Gemma-3 Family}
We use two instruction-tuned models from the Gemma-3 family: Gemma-3-4B-Instruct and Gemma-3-12B-Instruct. Gemma-3 models are trained with different data and optimization strategies compared to LLaMA-3 and Qwen-3, offering an additional axis of architectural and training diversity. We employ pretrained sparse autoencoders from GemmaScope for both Gemma-3 models. Results on Gemma-3 help assess the robustness of latent steering across independently developed model families.

\subsection{Input Prompts}
\label{sec:input_prompts}

We use fixed prompt templates for all experiments to ensure comparability across models and conditions. For each model family, we define a \textit{direct} prompt and a \textit{chain-of-thought (CoT)} prompt that differ only in whether explicit step-by-step reasoning is encouraged. Unless otherwise specified, all prompts are applied consistently across datasets, with decoding parameters held fixed. Due to the length of the prompt templates and the presence of model-specific special tokens, we present the full prompts as figures for readability.

\subsubsection{LLaMA-3 Family}
\begin{figure}[H]
    \centering
    \includegraphics[width=\linewidth]{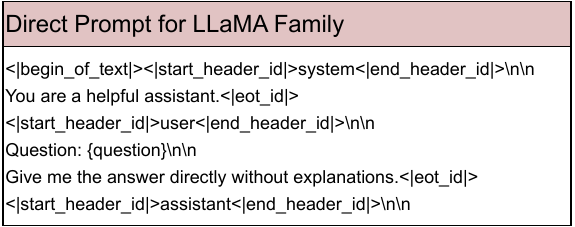}
    \caption{Direct prompt template for the LLaMA-3 family.}
    \label{fig:prompt_llama_direct}
\end{figure}

\begin{figure}[H]
    \centering
    \includegraphics[width=\linewidth]{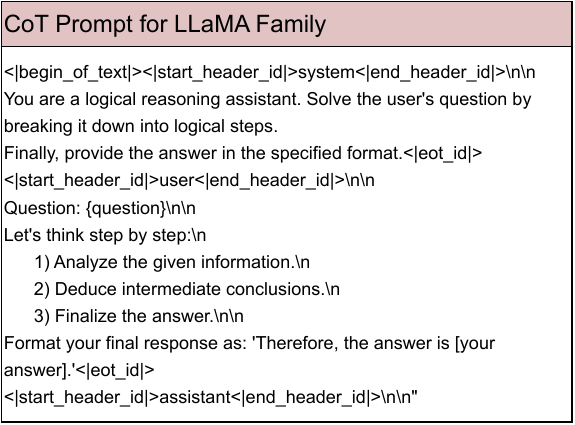}
    \caption{Chain-of-thought prompt template for the LLaMA-3 family.}
    \label{fig:prompt_llama_cot}
\end{figure}

For models in the LLaMA-3 family, we use fixed direct and chain-of-thought prompt templates.
The full prompt templates are shown in Figures~\ref{fig:prompt_llama_direct} and~\ref{fig:prompt_llama_cot}.

\begin{figure}[H]
    \centering
    \includegraphics[width=\linewidth]{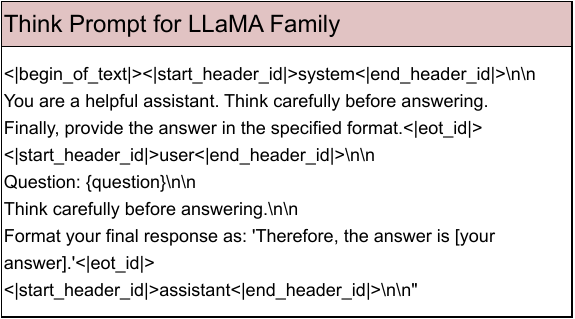}
    \caption{Think prompt template for the LLaMA-3 family.}
    \label{fig:prompt_llama_think}
\end{figure}

\begin{figure}[H]
    \centering
    \includegraphics[width=\linewidth]{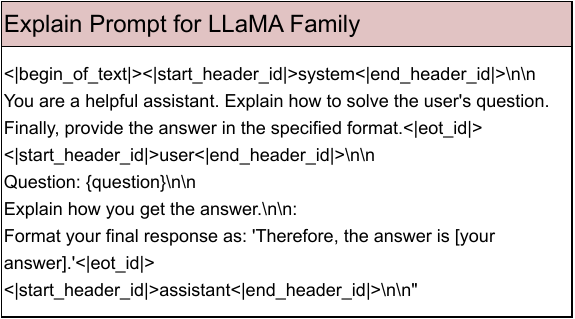}
    \caption{Explain prompt template for the LLaMA-3 family.}
    \label{fig:prompt_llama_explain}
\end{figure}

\begin{figure}[H]
    \centering
    \includegraphics[width=\linewidth]{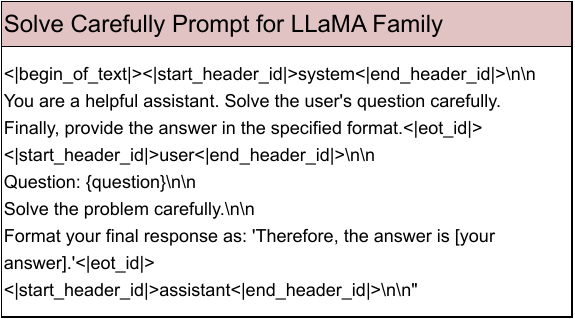}
    \caption{Solve carefully prompt template for the LLaMA-3 family.}
    \label{fig:prompt_llama_solve}
\end{figure}

In addition to the direct and CoT prompts, we also consider several alternative prompts for the cross-prompt analysis in Section~\ref{sec:cross_prompt}. These prompts implicitly encourage careful reasoning without explicitly requesting step-by-step explanations. The full templates for these prompts are shown in Figures~\ref{fig:prompt_llama_think}--\ref{fig:prompt_llama_solve}.

\paragraph{Think Prompt.}
See Figure~\ref{fig:prompt_llama_think} for the full template.

\paragraph{Explain Prompt.}
See Figure~\ref{fig:prompt_llama_explain} for the full template.

\paragraph{Solve Carefully Prompt.}
See Figure~\ref{fig:prompt_llama_solve} for the full template.

\subsubsection{Qwen-3 Family}
\begin{figure}[H]
    \centering
    \includegraphics[width=\linewidth]{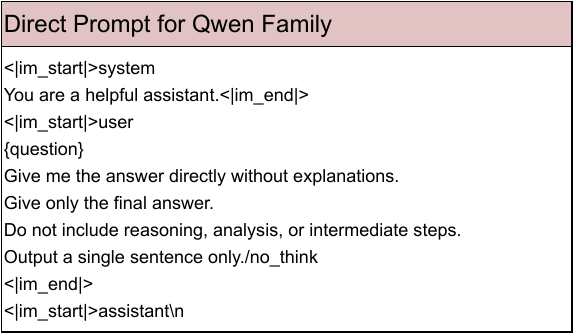}
    \caption{Direct prompt template for the Qwen-3 family.}
    \label{fig:prompt_qwen_direct}
\end{figure}

\begin{figure}[H]
    \centering
    \includegraphics[width=\linewidth]{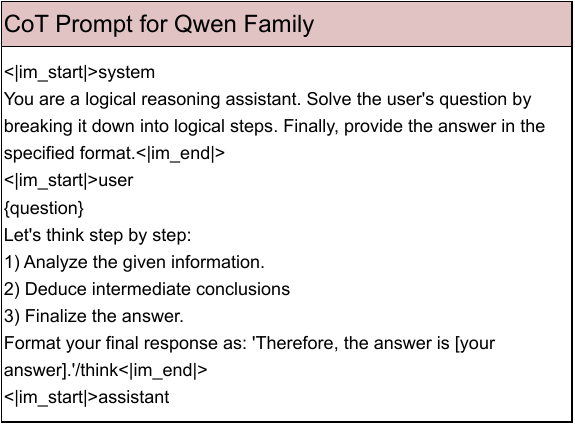}
    \caption{Chain-of-thought prompt template for the Qwen-3 family.}
    \label{fig:prompt_qwen_cot}
\end{figure}

For the Qwen-3 family, we define direct and chain-of-thought prompts analogous to those used for LLaMA-3, adapted to the instruction format expected by Qwen models. The full prompt templates are shown in Figures~\ref{fig:prompt_qwen_direct} and~\ref{fig:prompt_qwen_cot}.

\paragraph{Direct Prompt.}
See Figure~\ref{fig:prompt_qwen_direct} for the full template.

\paragraph{Chain-of-Thought Prompt.}
See Figure~\ref{fig:prompt_qwen_cot} for the full template.

\subsubsection{Gemma-3 Family}
\begin{figure}[H]
    \centering
    \includegraphics[width=\linewidth]{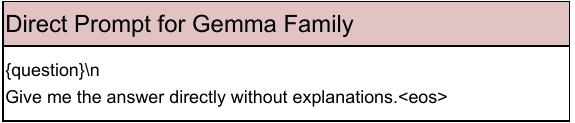}
    \caption{Direct prompt template for the Gemma-3 family.}
    \label{fig:prompt_gemma_direct}
\end{figure}

\begin{figure}[H]
    \centering
    \includegraphics[width=\linewidth]{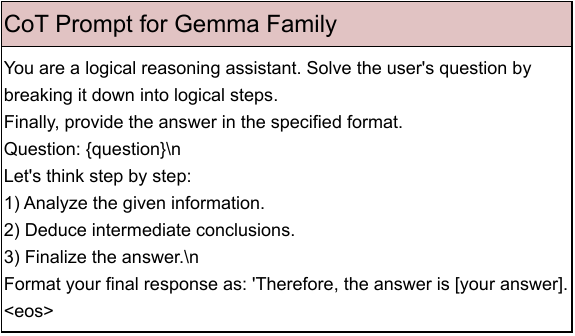}
    \caption{Chain-of-thought prompt template for the Gemma-3 family.}
    \label{fig:prompt_gemma_cot}
\end{figure}

For the Gemma-3 family, we likewise use a direct prompt and a chain-of-thought prompt that follow the instruction conventions of Gemma models. The full templates are shown in Figures~\ref{fig:prompt_gemma_direct} and~\ref{fig:prompt_gemma_cot}.

\paragraph{Direct Prompt.}
See Figure~\ref{fig:prompt_gemma_direct} for the full template.

\paragraph{Chain-of-Thought Prompt.}
See Figure~\ref{fig:prompt_gemma_cot} for the full template.

\begin{table*}[t]
\centering
\resizebox{\textwidth}{!}{
\begin{tabular}{lcccc}
\toprule
\textbf{Model} & \textbf{Steered Feature Index} & \textbf{Steering Step} & \textbf{Steering Strength $\alpha$} & \textbf{Steering Layer}\\
\midrule

\makecell[l]{LLaMA-3.1-8B-Instruct\\ \small \cite{meta_llama3_2024}}
&\#8629  & 1 &15&19  \\
\midrule

\makecell[l]{LLaMA-3.3-70B-Instruct\\ \small \cite{llama3.3-70b-instruct}}
& \#13709 & 1 &20&50  \\
\midrule

\makecell[l]{Qwen3-0.6B\\ \small \cite{qwen3technicalreport}}
& \makecell[c]{\#15352, \#14424, \#27058, \#22718, \#12509}
 & 1 & 25&19 \\
\midrule

\makecell[l]{Qwen3-4B\\ \small \cite{qwen3technicalreport}}
& \makecell[c]{\#12714, \#66391, \#58281, \#60128, \#25787}
 & 1 & 70&29 \\
\midrule

\makecell[l]{Gemma-3-4B-Instruct\\ \small \cite{gemma_2025}}
&  \makecell[c]{\#113617, \#80327, \#165330, \#52856, \#923, \\
\#51564, \#8253, \#6879, \#1257, \#5239}
& 1 & 15&29 \\
\midrule

\makecell[l]{Gemma-3-12B-Instruct\\ \small \cite{gemma_2025}}
& \makecell[c]{\#6662, \#7847, \#97, \#15521,  \#5100, \\
\#32828,  \#2616,  \#3397, \#12059, \#130}& 1 & 25&31  \\
\bottomrule
\end{tabular}
}
\caption{
Steering configurations for different models.
For each model, we report the index of the steered reasoning-related feature, the decoding step at which steering is applied, the steering strength $\alpha$, and the steering layer.
}
\label{tab:steering_settings}
\end{table*}

\paragraph{Prompt Consistency.}
Across all model families, the direct and CoT prompts differ only in their instructions regarding explicit reasoning. No task-specific information or answer hints are introduced through prompt wording. This design allows us to attribute observed differences in behavior to latent steering and prompting conditions rather than to prompt-specific artifacts.


\subsection{SAE Training Information}
\label{sec:train_sae}
For the Qwen3 models, we train sparse autoencoders (SAEs) from scratch using the \texttt{sparsify} library with its default configuration.
All training hyperparameters follow the library defaults unless otherwise specified.

We use the \texttt{EleutherAI/SmolLM2-135M-10B} dataset as training data for the SAEs. This dataset is a sampled subset of the SmolLM2-135M pretraining corpus (derived from the data used to train SmolLM2 as described in \cite{ben2025smollm2}), intended for efficient training of auxiliary components.
The dataset is tokenized using the corresponding model tokenizer and processed into fixed-length chunks following the standard \texttt{chunk\_and\_tokenize} procedure provided by the library.

SAEs are trained on the frozen base language model in an inference-only setting.
We use a batch size of 16 and train the autoencoders using the default optimization and sparsity settings.
No model-specific tuning is performed beyond selecting the target layers corresponding to the Qwen3 architectures.
For Qwen3-4B and Qwen3-0.6B, we train sparse autoencoders with latent dimensions of 26{,}624 and $32\times$ the model hidden size, respectively.
Both models use Top-$K$ sparsity with $K=192$.

\subsection{Steering Settings}
\label{sec:steering_settings}
Table~\ref{tab:steering_settings} summarizes the steering configurations used across different models.
For each model, we report the index of the reasoning-related feature selected for steering, the decoding step at which the intervention is applied, and the steering strength $\alpha$.
These settings are held fixed across tasks and datasets for a given model, and are used consistently in all steering experiments reported in the paper.

\subsection{Model-specific Configurations}
\label{sec:config}

All experiments use pretrained causal language models loaded from their official HuggingFace checkpoints, without any additional finetuning.
Models are evaluated in inference mode with parameters frozen.
We enable the output of hidden states (\texttt{output\_hidden\_states=True}) for all models to support activation-level analysis.
Models are loaded using standard half-precision (FP16), except for Gemma models which are evaluated in full precision (FP32).
Tokenizers use left padding, with the end-of-sequence token assigned as the padding token when necessary.


\subsection{Dense Steering Baseline}
\label{app:dense}

To compare SAE feature steering with a standard representation engineering baseline, we construct a dense steering direction based on the activation difference between CoT and direct prompting.

\paragraph{Dense Direction Construction.}
Given a set of problems from the GSM8K training set, we construct two prompts for each problem: a direct prompt and a CoT prompt (e.g., ``Let's think step by step''). We extract the hidden state of the last prompt token at the same transformer layer used for SAE feature discovery.

Let $h^{(i)}_{\text{cot}}$ and $h^{(i)}_{\text{direct}}$ denote the hidden states for the $i$-th example under CoT and direct prompting. The dense steering direction is computed as the mean activation difference:

\[
v_{\text{cot}} = \mathbb{E}_i[h^{(i)}_{\text{cot}}] - \mathbb{E}_i[h^{(i)}_{\text{direct}}].
\]

\paragraph{Steering Procedure.}
During generation under direct prompting, the dense direction is added to the hidden state of the same layer:

\[
h' = h + \alpha v_{\text{cot}},
\]

where $\alpha$ controls the steering strength. The intervention is applied at the first decoding step, consistent with the SAE steering experiments.

\paragraph{Evaluation Setup.}
The dense direction is constructed using the GSM8K training set only. The resulting direction is then applied to GSM8K test and BBH without modification. For each model, the steering strength $\alpha$ is tuned on a validation split to achieve the best reasoning performance.

\begin{table}[h]
\centering
\begin{tabular}{lc}
\toprule
Model & Steering Strength $\alpha$ \\
\midrule
LLaMA-3.1-8B-Instruct & 10 \\
Qwen3-4B & 10 \\
Gemma-3-4B-Instruct & 10 \\
\bottomrule
\end{tabular}
\caption{Steering strength used for dense representation steering.}
\label{tab:dense_alpha}
\end{table}

\subsection{Evaluation on a Large Reasoning Model}
\label{sec:lrm}

We additionally evaluate our method on a large reasoning model,
DeepSeek-R1-Distill-Llama-8B~\cite{deepseekai2025deepseekr1incentivizingreasoningcapability}, to examine whether the identified reasoning-related features remain effective in a model explicitly trained for long-form reasoning. We use the publicly available SAE checkpoint \texttt{andreuka18/sae-deepseek-r1-llama-8b}, using the SAE at layer 19.
To construct the direct-answer setting, we prepend an empty thinking block, \texttt{<think>\textbackslash n\textbackslash n</think>}, before generation, which suppresses explicit reasoning and encourages the model to produce a direct answer.

Table~\ref{tab:lrm_results} reports results on GSM8K, GPQA, and BBH. Across all three datasets, latent steering substantially improves the direct-answer baseline. On GSM8K, accuracy increases from 36.8\% to
74.8\%, approaching the 77.4\% accuracy obtained with standard CoT generation. Similarly, on BBH, steering improves accuracy from 41.6\% to 96.4\%, compared with 98.4\% under CoT. These results suggest that the steering effect extends to an R1-distilled reasoning model.

GPQA exhibits a somewhat different behavior. Steering improves the direct-answer accuracy from 34.3\% to 41.4\%, approaching the 42.9\% accuracy of CoT generation. However, both steered direct and CoT generation require substantially longer outputs on GPQA, exceeding 3,500 generated tokens on average. This indicates that, for the R1-distilled model, improved GPQA performance is accompanied by substantially increased generation cost.

\begin{table}[h]
\centering
\small
\begin{tabular}{llcc}
\toprule
\textbf{Dataset} & \textbf{Strategy}
& \textbf{Acc. (\%)}
& \textbf{Avg. Tokens} \\
\midrule
GSM8K
    & Direct          & 36.8 & 66   \\
    & Steered Direct  & 74.8 & 651  \\
    & CoT             & 77.4 & 694  \\
\midrule
GPQA
    & Direct          & 34.3 & 105  \\
    & Steered Direct  & 41.4 & 3742 \\
    & CoT             & 42.9 & 3573 \\
\midrule
BBH
    & Direct          & 41.6 & 26   \\
    & Steered Direct  & 96.4 & 294  \\
    & CoT             & 98.4 & 495  \\
\bottomrule
\end{tabular}
\caption{
Evaluation of latent steering on
DeepSeek-R1-Distill-Llama-8B. Steering consistently improves the
direct-answer baseline across GSM8K, GPQA, and BBH, approaching
standard CoT accuracy on all three benchmarks.
}
\label{tab:lrm_results}
\end{table}

\subsection{Case Study: First-Token Distribution Shift}

\begin{figure}[H]
\centering
\includegraphics[width=0.9\linewidth]{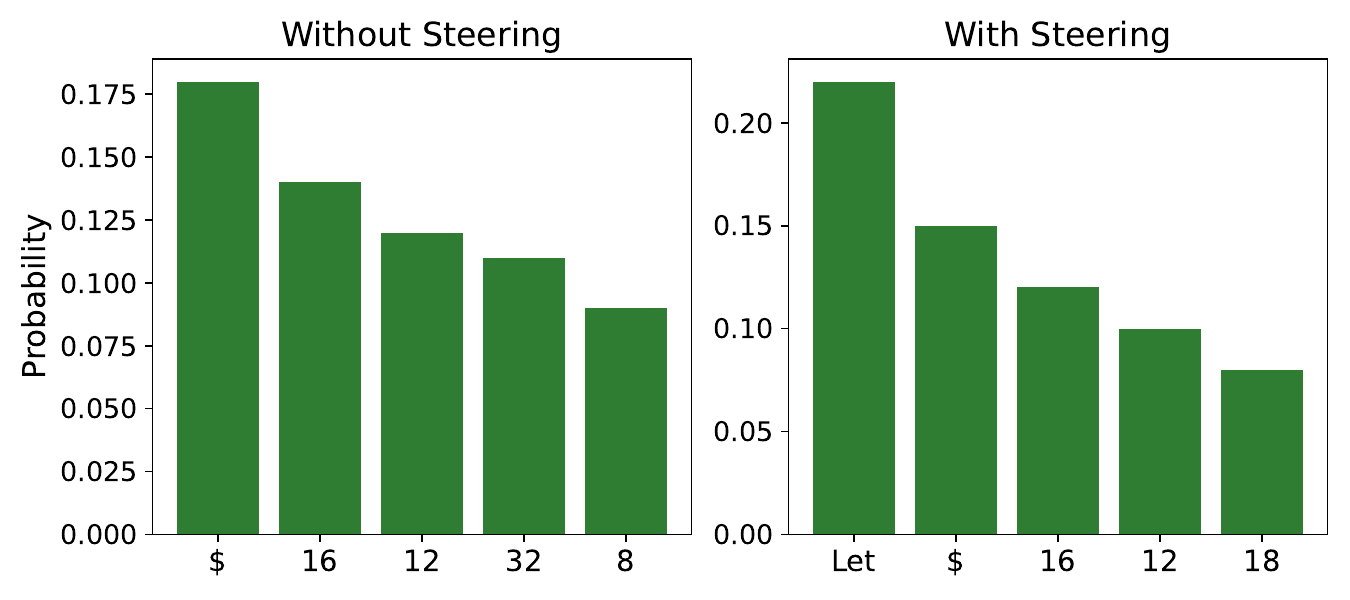}
\caption{
First-token probability distribution with and without steering. 
}
\label{fig:first_token_shift}
\end{figure}

To illustrate how early steering alters the decoding trajectory, we examine the first-token probability distribution with and without steering on GSM8K using LLaMA-3.1-8B-Instruct. Figure~\ref{fig:first_token_shift} shows the top predicted tokens under both conditions. Without steering, the model's highest-probability tokens are dominated by answer-like tokens such as numbers or symbols (e.g., ``\$'', ``16'', ``12''), indicating a direct-answer generation pattern. In contrast, when steering is applied, the top prediction becomes ``Let'', a common token used to begin step-by-step explanations. This qualitative observation suggests that steering shifts the model from an answer-generation trajectory to a reasoning trajectory at the earliest decoding step.

\subsection{Effect of Intervention Timing}
\label{sec:steer_timing}
We analyze how the effectiveness of steering depends on the timing of the intervention during generation.
Following the feature ranking defined in Section~\ref{sec:feature_discovery}, we consider both an intervention on the top-ranked feature (feature \#8629 in LLaMA-3.1-8B-Instruct) and, for comparison, an intervention on the top-10 reasoning-related features.
In all cases, we apply the intervention at a single decoding step while varying the step index.
\begin{figure}[h]
    \centering
    \includegraphics[width=\linewidth]{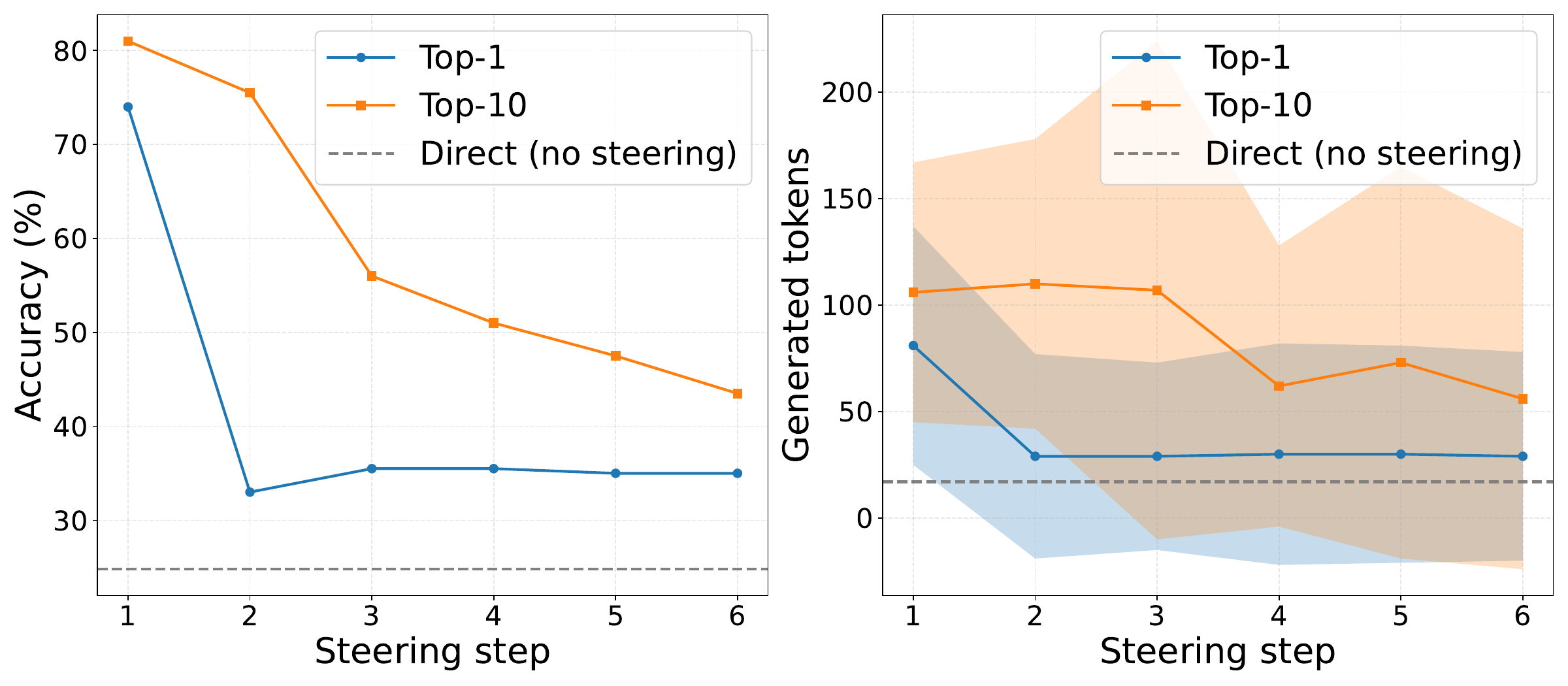}
    \caption{
    Effect of intervention timing when steering reasoning-related latent features on GSM8K with LLaMA-3.1-8B-Instruct.
    The figure reports accuracy (left) and generated token length (right) as a function of the decoding step at which the intervention is applied.
    }
    \label{fig:timing}
\end{figure}

As shown in Figure~\ref{fig:timing}, earlier interventions consistently yield stronger effects.
Intuitively, entering a reasoning mode early gives the model more opportunity to carry out the intermediate computations needed for a correct solution, whereas late interventions may occur when the model is already close to, or has effectively committed to, an answer.
This timing consideration is important for direct prompting, where some responses terminate within only a few tokens (e.g., $\sim$3 tokens), making later intervention steps ill-defined or ineffective.

Comparing top-1 and top-10 steering, both show a similar dependence on intervention timing, with earlier interventions being more effective, but their behaviors differ. As the intervention is delayed, top-10 steering weakens gradually, whereas top-1 steering drops sharply after the first step and quickly approaches the unsteered baseline. This suggests that a single feature can trigger reasoning when applied early, while sustained reasoning and its verbalization depend on multiple latent features, which is also reflected in the longer generations produced by top-10 steering.

\subsection{Reasoning Feature as a Mode Indicator}
\label{sec:reason_mode}
We analyze whether the identified feature is associated with \emph{entering a reasoning mode} or with \emph{reasoning quality}. 
Specifically, we examine the correlation between feature activation and (i) prompting strategy (CoT vs.\ direct), and (ii) answer correctness within each strategy.
Figure~\ref{fig:dynamic} shows a characteristic early activation pattern during answer generation.
We quantify this pattern by comparing the maximum activation within the first 20 decoding steps under CoT and direct prompting.

As shown in Table~\ref{tab:feature_correlation}, we find that the feature is activated significantly more strongly under CoT prompting than under direct prompting.
Across 200 GSM8K samples, the activation of the feature is substantially higher for CoT than for direct prompting (point-biserial $r=0.14$, $p=0.006$), indicating a strong association with entering a reasoning-oriented generation mode. Computation details are provided in~\ref{sec:correlation}.

\begin{table}[h]
\centering
\newcommand{\pmgray}[1]{\textcolor{gray}{\text{\scriptsize$\pm$#1}}}
\resizebox{\columnwidth}{!}{
\begin{tabular}{lcccc}
\toprule
\textbf{Target Variable} &\textbf{ corr. ($p$-value)}& \textbf{Mean (A)} &\textbf{ Mean (B)}  \\
\midrule
Reasoning Mode (A=CoT, B=Direct) & 0.14 (0.006)& 0.10\pmgray{0.41} & 0.01\pmgray{0.13}  \\
CoT Acc.  (A=Correct, B=Wrong)& $-0.02$ (0.80)& 0.10\pmgray{0.40} & 0.12\pmgray{0.43}  \\
Direct Acc. (A=Correct, B=Wrong) & $-0.06$ (0.40)& 0.00\pmgray{0.00} & 0.02\pmgray{0.13}  \\
\bottomrule
\end{tabular}}
\caption{
Analysis of the correlation between feature activation values and different target variables. 
Activation values are collected from feature \#8629 in LLaMA-3.1-8B-Instruct on GSM8K.
We report the correlation coefficients (with $p$-values) and the mean feature scores corresponding to different groups in the target variable.
}
\label{tab:feature_correlation}
\end{table}

In contrast, when conditioning on a fixed prompting strategy (either CoT or direct), feature activation shows little correlation with answer correctness.
As shown in Table~\ref{tab:feature_correlation}, the feature does not reliably distinguish correct from incorrect answers under either strategy.

This interpretation also explains the limited gains observed when steering the feature under CoT prompting in Table~\ref{tab:statistic_result}: \emph{once the model has already entered a reasoning mode, further amplifying the trigger alone does not substantially improve reasoning performance.}

\subsection{Point-biserial Correlation Details}
\label{sec:correlation}
We measure the association between the activation magnitude of a latent feature and a binary target variable using the point-biserial correlation. For each sample $i$, let $x_i$ denote the feature activation value (in our case, the maximum activation within the first 20 decoding steps), and let $y_i \in \{0,1\}$ denote the binary label indicating whether the sample belongs to condition A ($y_i=1$) or condition B ($y_i=0$).

Let $n_1$ be the number of samples with $y_i=1$ and $n_0$ the number of samples with $y_i=0$, with $n=n_1+n_0$. Define the group means
\[
\bar{x}_1 = \frac{1}{n_1}\sum_{i:y_i=1} x_i, \qquad
\bar{x}_0 = \frac{1}{n_0}\sum_{i:y_i=0} x_i,
\]
and the overall sample standard deviation of $x$
\[
s_x = \sqrt{\frac{1}{n-1}\sum_{i=1}^{n} (x_i-\bar{x})^2},
\qquad \text{where } \bar{x}=\frac{1}{n}\sum_{i=1}^n x_i.
\]
The point-biserial correlation coefficient is then
\[
r_{pb} = \frac{\bar{x}_1 - \bar{x}_0}{s_x}\sqrt{\frac{n_1 n_0}{n^2}}.
\]
(Equivalently, $r_{pb}$ is identical to the Pearson correlation between $x_i$ and the binary variable $y_i$ when $y_i$ is coded as $0/1$.)

To test significance under the null hypothesis $H_0: r_{pb}=0$, we use the standard $t$-test for correlation:
\[
t = r_{pb}\sqrt{\frac{n-2}{1-r_{pb}^2}},
\]
with degrees of freedom $n-2$. We report two-sided $p$-values computed from the $t$ distribution:
\[
p = 2\left(1 - F_{t(n-2)}\left(|t|\right)\right),
\]
where $F_{t(n-2)}(\cdot)$ denotes the CDF of the $t$ distribution with $n-2$ degrees of freedom.

\subsection{Ablation Study}

We conduct a random feature ablation to verify that the observed steering effects are specific to the identified reasoning-related latent feature rather than arising from arbitrary perturbations.
Using LLaMA-3.1-8B-Instruct under direct prompting, we compare steering the reasoning feature (\#8629) with steering randomly selected non-zero SAE features from the same layer.

\begin{table}[h]\small
\centering
\newcommand{\pmgray}[1]{\textcolor{gray}{\text{\scriptsize$\pm$#1}}}
\resizebox{0.8\columnwidth}{!}{
\begin{tabular}{l|ccc}
\toprule
Feature & Acc. (\%) & \#Tok & Runs \\
\midrule
\#8629 & 73.3 & 83 & - \\
Random   & $26.1 \pmgray{3.4} $ & $29 \pmgray{5}$ & 10 \\
\bottomrule
\end{tabular}}
\caption{Random feature ablation on GSM8K with LLaMA-3.1-8B-Instruct. Results for the reasoning-related feature are reported with a fixed random seed.}
\label{tab:random_ablation}
\end{table}

As shown in Table~\ref{tab:random_ablation}, steering the reasoning-related feature yields substantial gains in accuracy and induces longer generations, whereas random feature steering fails to produce comparable improvements.
This result confirms that the steering effect is specific to the identified reasoning-related feature rather than a generic latent intervention.




\end{document}